\documentclass[11pt]{article}

\usepackage[final]{acl}
\usepackage{times}
\usepackage{latexsym}
\usepackage[T1]{fontenc}
\usepackage[utf8]{inputenc}
\usepackage{microtype}
\usepackage{inconsolata}
\usepackage{graphicx}

\usepackage{wrapfig}
\usepackage{booktabs}
\usepackage{float}
\usepackage{xcolor}
\usepackage{colortbl}
\usepackage{multirow}

\definecolor{lightgreen}{rgb}{0, 0.6, 0.3}
\definecolor{lightred}{rgb}{1, 0.2, 0.2}
\definecolor{mygray}{gray}{0.9}

\usepackage{tcolorbox}
\tcbuselibrary{breakable,skins}
\definecolor{promptbgcolor}{gray}{0.95}
\definecolor{promptframecolor}{gray}{0.7}
\newtcolorbox{promptdisplay}[1]{
  skin=enhanced,
  colback=promptbgcolor,
  colframe=promptframecolor,
  coltitle=black,
  boxsep=5pt,
  left=5pt, right=5pt, top=5pt, bottom=5pt,
  arc=4pt,
  outer arc=4pt,
  fonttitle=\bfseries,
  title={#1},
  breakable,
}

\usepackage{amsmath,amsfonts,bm}

\def\eqref#1{equation~\ref{#1}}

\def\1{\bm{1}}

\def\vS{{\bm{S}}}

\DeclareMathAlphabet{\mathsfit}{\encodingdefault}{\sfdefault}{m}{sl}
\SetMathAlphabet{\mathsfit}{bold}{\encodingdefault}{\sfdefault}{bx}{n}

\newcommand{\squishlist}{
\begin{list}{{{\small{$\bullet$}}}}
{\setlength{\itemsep}{3pt}      \setlength{\parsep}{1pt}
\setlength{\topsep}{1pt}       \setlength{\partopsep}{0pt}
\setlength{\leftmargin}{1em} \setlength{\labelwidth}{1em}
\setlength{\labelsep}{0.5em} } }
\newcommand{\squishend}{  \end{list}  }

\title{Evian: Towards Explainable Visual Instruction-tuning Data Auditing}

\author{
  \textbf{Zimu Jia\textsuperscript{1}},
  \textbf{Mingjie Xu\textsuperscript{1}},
  \textbf{Andrew Estornell\textsuperscript{2}},
  \textbf{Jiaheng Wei\textsuperscript{1}\thanks{Corresponding author}} \\
  \\
  \textsuperscript{1}The Hong Kong University of Science and Technology (Guangzhou) \\
  \textsuperscript{2}ByteDance Seed \\
  \\
  \small{
    \texttt{jiahengwei@hkust-gz.edu}
  }
}

\begin{document}
\maketitle
\begin{abstract}
The efficacy of Large Vision-Language Models (LVLMs) is critically dependent on the quality of their training data, requiring a precise balance between visual fidelity and instruction-following capability. Existing datasets, however, are plagued by inconsistent quality, and current data filtering methods rely on coarse-grained scores that lack the granularity to identify nuanced semantic flaws like logical fallacies or factual errors. This creates a fundamental bottleneck in developing more reliable models. To address this, we make three core contributions. First, we construct a large-scale, 300K-sample benchmark by systematically injecting diverse, subtle defects to provide a challenging testbed for data auditing. Second, we introduce a novel ``Decomposition-then-Evaluation'' paradigm that breaks model responses into constituent cognitive components: visual description, subjective inference, and factual claim, enabling targeted analysis. Third, we instantiate this paradigm via \textbf{EVIAN} (\textbf{E}xplainable \textbf{V}isual \textbf{I}nstruction-tuning Data \textbf{A}uditi\textbf{N}g), a pipeline that evaluates these components along the orthogonal axes of Image-Text Consistency, Logical Coherence, and Factual Accuracy. Our empirical findings challenge the prevailing scale-centric paradigm: a model fine-tuned on a compact, high-quality subset curated by EVIAN consistently surpassed models trained on orders-of-magnitude larger datasets. We also reveal that dividing complex auditing into verifiable subtasks enables robust curation, and that Logical Coherence is the most critical factor in data quality evaluation.
\end{abstract}

\section{Introduction}
\label{sec:intro}
Large Vision-Language Models (LVLMs) \citep{chen2024dress} have recently demonstrated remarkable progress in aligning visual perception with natural language understanding, enabling a wide range of applications from medical assistance to robotic control \citep{yin2024survey,li2025recognition,pang2025vlms, yue2024mmal}. An important factor of this success is \textit{Visual Instruction Tuning} (VIT), which aligns visual representations with language instructions to enhance instruction-following capability \citep{liu2023visual}. However, the effectiveness of VIT hinges on the quality of the underlying training data, which must strike a delicate balance between adhering to user commands and maintaining fidelity to visual inputs. 

Existing datasets and filtering methods fall short of this requirement. Large-scale data synthesis (e.g., LLaVA-Instruct-150K) improves instruction following but often introduces noise \citep{liu2024mminstruct, tang2024textsquare}, while similarity-based filtering methods (e.g., CLIP score) promote visual grounding but lack the granularity to detect subtle semantic flaws \citep{wang2024vigc}. As a result, current LVLMs frequently suffer from fine-grained errors, including object hallucination, attribute misattribution, factual inconsistency, and flawed reasoning \citep{liu2024survey, bai2024hallucination, chen2024multi}. These deficiencies reveal a fundamental bottleneck: prevailing approaches rely on coarse, uni-dimensional quality measures that collapse diverse error types into a single opaque score.

In this work, we argue that evaluating model-generated responses requires moving beyond monolithic scoring toward structured verification. Our core insight is that a response is not an indivisible block of text but a composite of distinct, verifiable components. Building on this principle, we propose the \textit{Decomposition-then-Evaluation} paradigm, which reframes the task of auditing complex responses into targeted sub-tasks. Specifically, we isolate and validate \textit{pure visual descriptions} to address visual misrepresentation, \textit{external factual claims} to correct factual inaccuracies, and \textit{subjective inferences} to mitigate flawed reasoning.

To operationalize this paradigm, we introduce \textbf{EVIAN} (\textbf{E}xplainable \textbf{V}isual \textbf{I}nstruction-tuning Data \textbf{A}uditi\textbf{N}g), an automated and interpretable framework that systematically evaluates responses along three orthogonal axes: Image-Text Consistency, Logical Coherence, and Factual Accuracy. Complementing this framework, we construct a large-scale, 300K-sample benchmark by injecting diverse, subtle defects, providing a challenging testbed for fine-grained data auditing. Our empirical findings show that models fine-tuned on compact, high-quality subsets curated by EVIAN consistently outperform models trained on orders-of-magnitude larger datasets, highlighting that interpretable data curation, rather than sheer scale, is the key to advancing LVLMs.

Our main contributions are as follows:  
\squishlist
    \item To spur research in LVLM visual instruction tuning data quality and facilitate rigorous evaluation, we introduce a 300K-sample benchmark for visual instruction data selection, built by systematically injecting diverse semantic defects to support fine-grained auditing.  
    \item We propose the \textit{Decomposition-then-Evaluation} paradigm and instantiate it in \textbf{EVIAN}, a fully automated and interpretable framework that decomposes responses into visual descriptions, subjective inferences, and factual claims, and evaluates them along three orthogonal dimensions.  
    \item We conduct extensive experiments showing that for LVLMs, the logical integrity of training data is a more decisive factor for downstream performance than its informational richness, establishing the critical need to prioritize reasoning and factual correctness in data curation. 
\squishend

\section{Related Work}
\label{sec:related}
Vision-language data curation has progressed from coarse pre-training filters to instruction-tuning strategies, yet scalable and fine-grained evaluation remains largely missing. As a result, most existing methods still rely on shallow quality proxies, limiting their ability to diagnose subtle semantic, logical, or factual defects.

\paragraph{Data Selection for Vision-Language Pre-training.}
A central challenge in vision-language learning is selecting high-quality subsets from noisy web-scale corpora such as LAION \citep{schuhmann2021400m}. Early approaches rely on similarity-based filtering with pre-trained models, including CLIP \citep{radford2021learning}, ALBEF \citep{li2021align}, and BLIP \citep{li2022blip, li2023blip}, using holistic similarity scores \citep{hessel2021clipscore, xu2025quality, wang2024cliploss} or mixture modeling \citep{shi2024breaking}. More recent work adopts fine-tuned multimodal language models as learned data filters, scoring image-text pairs along multiple semantic dimensions \citep{wang2024finetuned}, or employs generative models for dataset sanitization via re-captioning and label correction \citep{vasa2025autovdc, mahjourian2025sanitizing, zhang2024data, zhu2023unmasking, zhang2025trustclip}. Despite their effectiveness, these methods predominantly depend on coarse proxies and offer limited insight into complex reasoning or factual errors.

\paragraph{Data Curation for Visual Instruction Tuning.}
As vision-language models shift from representation learning to instruction following \citep{safaei2025filter}, data quality has become increasingly critical \citep{chen2024your}. The ``quality over quantity'' principle was first demonstrated in the text domain by AlpaGasus \citep{chen2023alpagasus} and later extended to multimodal settings by InstructionGPT-4 \citep{wei2023instructiongpt}, which combines CLIP scores with GPT-4 judgments to curate compact datasets. Other approaches generate synthetic instruction data \citep{liu2024synthvlm, chen2024spatialvlm} or augment supervision with reasoning traces, such as Reflective Instruction Tuning \citep{zhang2024reflective}. A widely adopted alternative is the LLM-as-a-Judge paradigm \citep{gu2024survey, li2024llms, pu2025judge}, which has been shown to suffer from bias, instability, and reasoning shortcuts, particularly without ground-truth references \citep{shi2024judging, hwang2025fooling, ye2024justice, guerdan2025validating, wei2024measuring, zhang2026rep}. Consequently, scalable and reliable instruction-tuning data curation remains an open problem.

\paragraph{The Gap in Fine-Grained Evaluation.}
The reliance on coarse filtering reflects a broader absence of scalable fine-grained evaluation. While prior work has explored alternatives to single holistic scores \citep{adlakha2024evaluating}, systematic diagnosis of semantic errors remains limited. Early automated methods rely on fixed criteria that struggle with open-ended errors \citep{zhao2024automated}, and recent pipelines such as SCALE \citep{xu2025better} lack explicit modeling of compositional reasoning. Task-specific benchmarks for logical reasoning \citep{xiao2024logicvista, xu2025visulogic} provide deeper analysis but are too narrow for general data auditing, while conceptual discussions of holistic evaluation \citep{tu2025toward} stop short of actionable frameworks. This gap motivates structured, component-level evaluation that disentangles visual grounding, reasoning, and factual correctness in multimodal data.

\section{The Method: Evian}
\label{sec:method}
\begin{figure*}[h]
    \centering
    \includegraphics[width=0.85\linewidth]{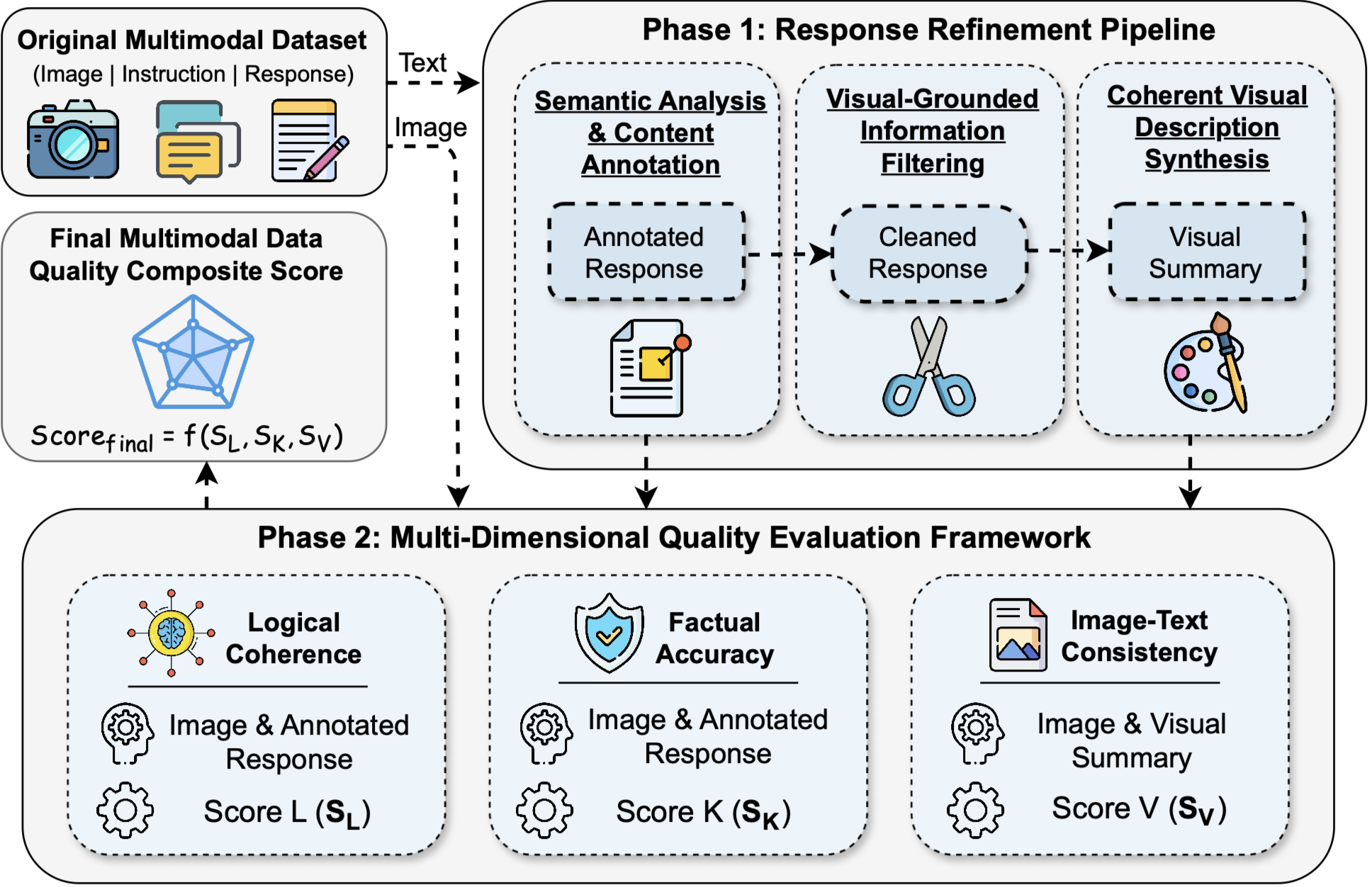}
    \caption{Overview of the two-phase EVIAN framework, which first decomposes a response into visual, inferential, and factual components and then evaluates them along the orthogonal dimensions of Image--Text Consistency, Logical Coherence, and Factual Accuracy.}
    \label{fig:overview}
\end{figure*}

We propose \textbf{EVIAN}, an automated pipeline for auditing visual instruction data. As illustrated in Figure~\ref{fig:overview}, EVIAN follows a two-phase process: (i) response decomposition, which disentangles complex answers into verifiable components, and (ii) multi-faceted evaluation, which scores these components across orthogonal quality dimensions.

\subsection{Problem Definition and Data Quality Metrics}
\label{sec:problem_definition}
We define \textit{visual instruction data auditing} as the task of assigning interpretable quality scores to image-instruction-response triples. Formally, given $x_i = (I_i, P_i, R_i)$ from dataset $D$, our auditing function $\Phi$ maps each sample to a three-dimensional score vector:
\begin{equation}
    \vS_i = \Phi(x_i) = (S_{L,i}, S_{K,i}, S_{V,i}),
\end{equation}
where each score ranges from 1 (low) to 5 (high). The three metrics are:  
\squishlist
    \item  \textbf{Logical Coherence ($S_L$)}: soundness of reasoning relative to the instruction and visual evidence.  
    \item  \textbf{Factual Accuracy ($S_K$)}: correctness of knowledge claims against external facts.  
    \item  \textbf{Image-Text Consistency ($S_V$)}: fidelity of the textual response to the visual input.  
\squishend

Together, these axes provide a comprehensive measure of data quality, capturing both semantic integrity and visual fidelity.

\subsection{Phase 1: Response Decomposition via Chain-of-Thought}
\label{sec:phase1_cot}
The first phase disentangles raw responses into verifiable components, separating visual descriptions from subjective inferences and factual claims. This is achieved through a three-step chain-of-thought (CoT) process, $\Psi_{\text{deconstruct}}$, implemented with the Qwen3-235B-A22B-Instruct-2507-FP8 model \citep{qwen3technicalreport}. The result is an annotated response with explicit tags and a purified visual summary, which together form the basis for systematic auditing.

\begin{figure}[H]
    \centering
    \includegraphics[width=1\linewidth]{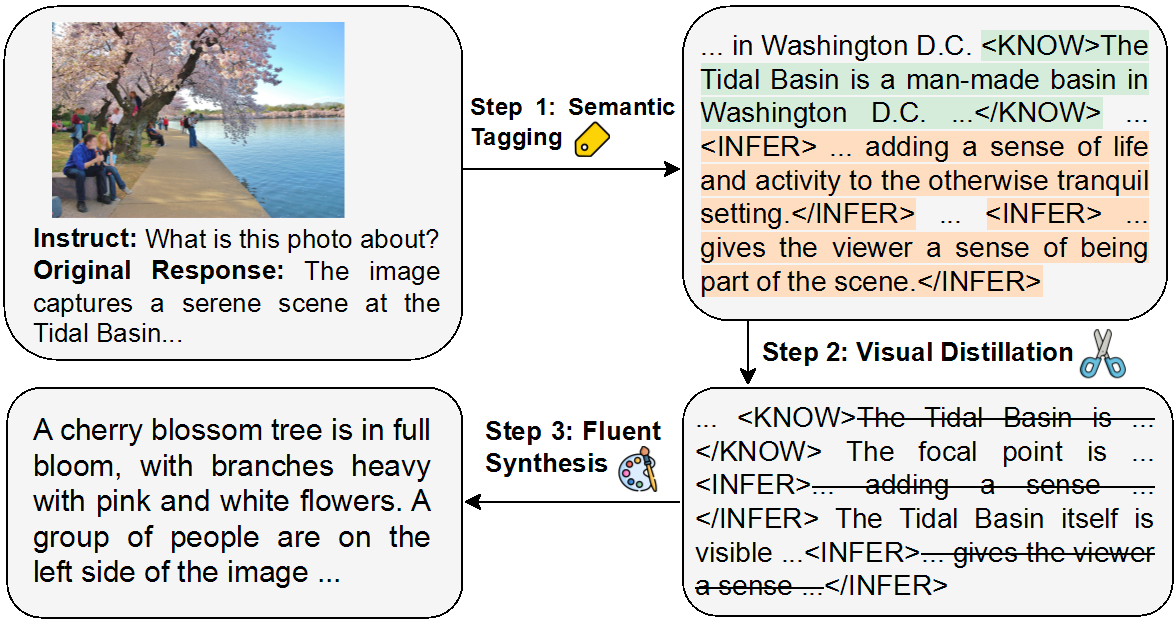}
    \caption{Three-stage Chain-of-Thought (CoT) process for response decomposition, which (1) isolates subjective inferences and factual claims via semantic tagging, (2) purifies the text through visual distillation, and (3) refines the output into a cohesive, purely visual summary.}
    \label{fig:CoT}
\end{figure}

\paragraph{Step 1: Semantic Tagging.}  
The process begins by parsing the raw response $R_i$ while strictly preserving its original wording. Subjective judgments (e.g., ``the room feels cozy'') are wrapped in \texttt{<INFER>} tags, and knowledge-dependent claims (e.g., ``this is a Bauhaus-style lamp'') are wrapped in \texttt{<KNOW>} tags. Untagged text is treated as purely visual description. This produces an annotated response $R_i^{\text{annotated}}$ that explicitly separates cognitive components without altering their content.

\paragraph{Step 2: Visual Distillation.}  
Next, the annotated response is distilled into a purely visual form. Segments within \texttt{<INFER>} or \texttt{<KNOW>} tags are either rewritten into neutral, descriptive statements or deleted if unverifiable. For example, ``this is likely a wedding dress'' becomes ``a white dress''; unverifiable claims are dropped entirely. Untagged visual statements remain unchanged. The result is a draft $R_i^{\text{draft}}$ containing only objective, image-grounded content.

\paragraph{Step 3: Fluent Synthesis.}  
Since distillation may fragment the text, a final synthesis step restores fluency and coherence. The draft response is reorganized into a single, natural paragraph while strictly forbidden from adding new content. This ensures the output $R_i^{\text{visual}}$ is a faithful, high-quality visual summary.  

Together, these steps yield two complementary artifacts: $R_i^{\text{annotated}}$, which retains the full response structure with explicit tags, and $R_i^{\text{visual}}$, which isolates objective descriptions. This decomposition provides the foundation for precise, component-level auditing in Phase 2.

\subsection{Phase 2: Multi-faceted Quality Assessment}
\label{sec:phase2_assessment}
The second phase conducts a multi-faceted evaluation of each decomposed response along three orthogonal dimensions: logical coherence, factual accuracy, and image-text consistency. We employ Qwen2.5-VL-7B-Instruct-AWQ \citep{bai2025qwen2} as an automated auditor, which assigns interpretable 1–5 scores and textual rationales based on a detailed rubric. This step provides fine-grained diagnostics of different error types while producing standardized quality scores that can be aggregated for ranking and selection.

\paragraph{Logical Coherence ($S_L$).}  
This dimension evaluates whether reasoning in the \texttt{<INFER>} tags follows plausibly from visual evidence. Scores increase with reasoning strength: a default of 2 when no inference is given, 3 for plausible but unsubstantiated claims, 4 for well-supported reasoning, and 5 for logically undeniable conclusions. This rubric rewards depth of reasoning while penalizing speculation.
\paragraph{Factual Accuracy ($S_K$).}  
This dimension fact-checks knowledge claims in the \texttt{<KNOW>} tags against the auditor’s internal knowledge. Fully correct claims receive 5, minor inaccuracies lower the score to 4, and a single major error (e.g., misidentifying a capital city) caps the score at 2. In the absence of knowledge claims, the default score is 2, distinguishing informative from non-informative responses.
\paragraph{Image-Text Consistency ($S_V$).}  
This dimension measures the alignment of the purified visual description $R^{\text{visual}}$ with the image. The principle is consistency over completeness: omissions are acceptable, but contradictions or unverifiable assertions are heavily penalized. Perfectly faithful descriptions receive 5, minor imprecisions result in 4, and any clear contradiction drops the score to 2 or below. This ensures that only visually accurate responses achieve the highest marks.

By producing a triplet $(S_L, S_K, S_V)$ with explicit explanations, Phase 2 delivers an interpretable and multi-dimensional quality assessment. These scores directly guide downstream data ranking and selection.

\subsection{Data Ranking and Selection}
\label{sec:ranking_selection}
To enable downstream filtering, the three-dimensional score vector $\vS$ is aggregated into a single scalar:
\begin{equation}
    S_{\text{overall}} = \frac{S_L + S_K + S_V}{3}.
\end{equation}
This default scheme assumes equal importance, but weights can be tuned for specific applications (e.g., emphasizing $S_K$ for knowledge-intensive tasks or $S_L$/$S_V$ for creative captioning). To investigate the impact of these variations, we conduct a sensitivity analysis on the component weights, which is detailed in Appendix~\ref{app:sensitivity}. This flexibility ensures that data selected by EVIAN aligns with diverse modeling objectives.

\section{Benchmarking Data Quality via Controlled Defect Injection}
\label{sec:benchmark_construction}
To quantitatively validate a data auditing pipeline’s ability to detect fine-grained flaws in logical coherence, factual accuracy, and image-text consistency, a tailored benchmark with systematically injected defects is essential, as existing datasets lack the controlled errors needed for such a targeted evaluation. To ensure consistency with prior work, we adopt the SCALE methodology \citep{xu2025better} as the starting point for benchmark construction. From its source pool of 500,000 multimodal samples across eight datasets (Table~\ref{tab:datasets}), we derive two complementary components: (i) a 50,000-sample “gold standard” set purified by SCALE, and (ii) a 250,000-sample “challenge” set obtained via random down-sampling followed by our defect injection pipeline. Together, these components yield a reproducible benchmark of 300,000 samples, designed to evaluate whether data auditing methods can distinguish clean data from semantically corrupted examples.

\begin{table}[h]
\caption{Overview of the source datasets, comprising 300K samples from eight foundational datasets grouped into General Vision-Language tasks and Domain-Specific Reasoning tasks.}
\label{tab:datasets}
\centering
\resizebox{\linewidth}{!}{
\begin{tabular}{@{}ll@{}}
\toprule
\textbf{Dataset} & \textbf{Task Category} \\ 
\midrule
\multicolumn{2}{l}{\cellcolor{mygray}\textit{\textbf{General Visual-Language Capabilities}}} \\ 
ShareGPT-4V \citep{chen2024sharegpt4v} & Instruction Following \\
LLaVA-1.5-Mix \citep{liu2024improved} & General QA \\
AllSeeing-V2 \citep{wang2024all} & Grounding \\ 
\addlinespace[3pt]

\multicolumn{2}{l}{\cellcolor{mygray}\textit{\textbf{Domain-Specific Reasoning Capabilities}}} \\ 
DocVQA \citep{mathew2021docvqa} & Document \\
ChartQA \citep{masry2022chartqa} & Chart \\
InfoVQA \citep{mathew2022infographicvqa} & OCR \\
A-OKVQA \citep{schwenk2022okvqa} & Knowledge \\ 
Geometry3K \citep{lu2021inter} & Mathematics \\
\bottomrule
\end{tabular}%
}
\end{table}

\begin{table*}[h] 
\caption{Principled taxonomy of semantic defects used in benchmark construction, categorizing errors into three dimensions aligned with EVIAN’s evaluation modules: \textbf{Consistency}, \textbf{Reasoning}, and \textbf{Knowledge}.
} 
\label{tab:error_catalog} 
\small
\centering
\begin{tabular}{@{}l p{0.65\textwidth}@{}} 
\toprule 
\textbf{Error Subtype} & \textbf{Description / Generation Strategy} \\ 
\midrule 

\multicolumn{2}{l}{\cellcolor{mygray}\textbf{\textit{Image-Text Consistency ($S_V$)}}} \\ 
Attribute & Describes an object's attribute (e.g., color, material) incorrectly. \\ 
Spatial & Details incorrect spatial relations between objects. \\ 
Action & Assigns a wrong action or state to a subject. \\ 
Fake & Introduces a plausible yet non-existent object. \\ 
Misidentification & Misidentifies an existing object. \\ 

\addlinespace[3pt] 

\multicolumn{2}{l}{\cellcolor{mygray}\textbf{\textit{Logical Coherence ($S_L$)}}} \\ 
Conclusion & Generalizes hastily from a single detail. \\ 
Causal & Mistakes correlation for causation between events. \\ 
Prediction & Makes a baseless prediction from scant evidence. \\ 
Procedural & Adds a flawed or superfluous step to a process. \\ 
Comparison & Forms a misleading analogy from superficial traits. \\ 

\addlinespace[3pt]

\multicolumn{2}{l}{\cellcolor{mygray}\textbf{\textit{Factual Accuracy ($S_K$)}}} \\ 
Entity & Corrupts facts about a named entity. \\ 
Context & Places an object in a wrong historical/technological context. \\ 
Definition & Provides an incorrect definition of a concept. \\ 
Attribution & Misattributes a quote or work to the wrong source. \\ 
\bottomrule 
\end{tabular} 
\end{table*}

\paragraph{Defect Injection Pipeline.}  
The challenge set is generated through a three-stage pipeline that leverages the Qwen3-235B-A22B-Instruct-2507-FP8 model \citep{qwen3technicalreport} to embed subtle, context-aware flaws. The process is guided by a principled taxonomy (Table~\ref{tab:error_catalog}) spanning three critical dimensions for auditing: \textit{perceptual consistency}, \textit{factual accuracy}, and \textit{logical coherence}.  

\textbf{Stage 1: Content Analysis.} Each source response is analyzed by an LLM to identify whether it contains external knowledge or logical reasoning. This structured analysis, output in JSON, serves as a prior to ensure that subsequent errors are coherent with the intrinsic properties of the text.  

\textbf{Stage 2: Contextual Error Selection.} An error category is chosen via a probabilistic cascade. To counter their rarity, knowledge-related and reasoning-related errors are prioritized with probabilities of 0.8 and 0.6, respectively, while perceptual consistency serves as the default. Subtypes are selected randomly for consistency errors, whereas an additional LLM call determines the most plausible subtype for knowledge and reasoning cases.  

\textbf{Stage 3: Guided Rewriting.} The chosen error is injected by prompting the LLM with a targeted transformation instruction. A strict system prompt constrains the model to output only the modified text, ensuring automation and reproducibility.  

This injection strategy goes beyond simple noise addition: it produces realistic, semantically rich corruptions aligned with the three audit dimensions. As a result, the benchmark offers a challenging testbed for assessing whether auditing pipelines can detect not only superficial inconsistencies but also deeper factual and logical flaws.

\section{Experiments}
\subsection{Experimental Setup}
\label{sec:experiments}
\paragraph{Baselines.}
We compare our method against a diverse set of data auditing baselines spanning visual-language pretraining filters and recent visual instruction tuning approaches. Specifically, we consider: (1) \textbf{Random Sampling}, which randomly selects 10,000 samples as a non-selective lower bound; (2) \textbf{Image-Text Similarity Filters}, including \textbf{CLIPScore} (ViT-B/32), \textbf{ALBEF}, \textbf{BLIP}, and \textbf{BLIP-2}, which rank the full data pool by holistic image-text similarity and select the top 10,000 samples; (3) \textbf{SCALE}, a multi-stage filtering method that evaluates modality quality, relevance, clarity, and task rarity using a weighted scoring scheme; and (4) \textbf{Qwen2.5-VL-7B-Instruct-AWQ}, which directly scores sample quality via model-based evaluation and selects the top 10,000 instances.

\begin{figure*}
    \centering
    \includegraphics[width=0.9\textwidth]{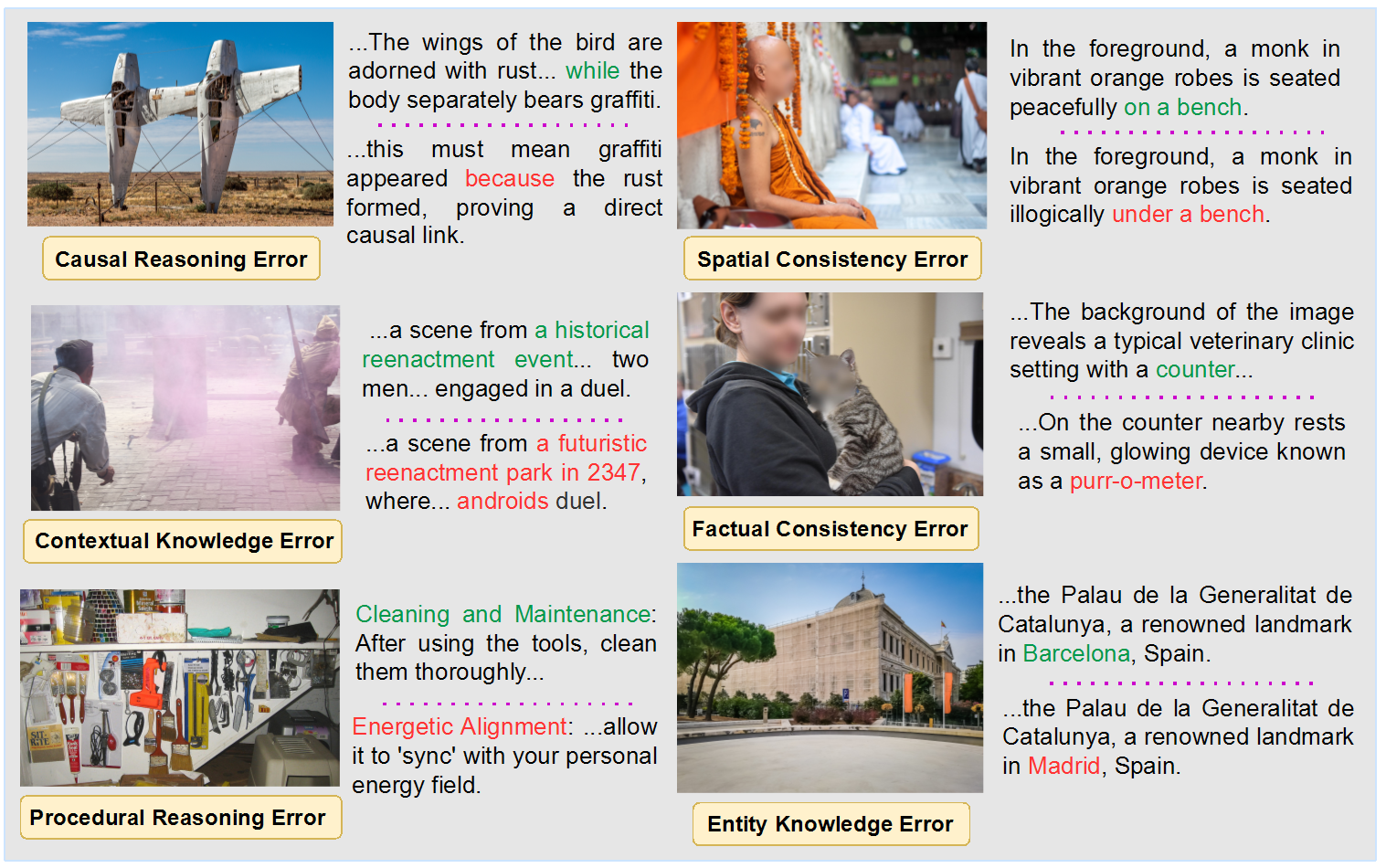}
    \caption{Examples of our controlled defect injection. For each pair, the \textcolor{lightgreen}{original high-quality text (top)} is rewritten to include a \textcolor{lightred}{subtle, context-aware flaw (bottom)}, illustrating various error categories from our taxonomy (Table~\ref{tab:error_catalog}).}
    \label{fig:defect_examples}
\end{figure*}

 \paragraph{Evaluation Protocol.}
For all methods, we fine-tune Qwen2-VL-2B on the selected 10,000-sample subset and evaluate the resulting models using VLMEvalKit \citep{duan2024vlmevalkit}. All experiments share identical architectures, SFT procedures, and hyperparameters, ensuring that performance differences reflect data quality rather than training variation.

\subsection{Evian Scores: Distribution and Discrimination}
\label{sec:score_distribution}
\begin{figure}[h]
    \centering
    \includegraphics[width=1\linewidth]{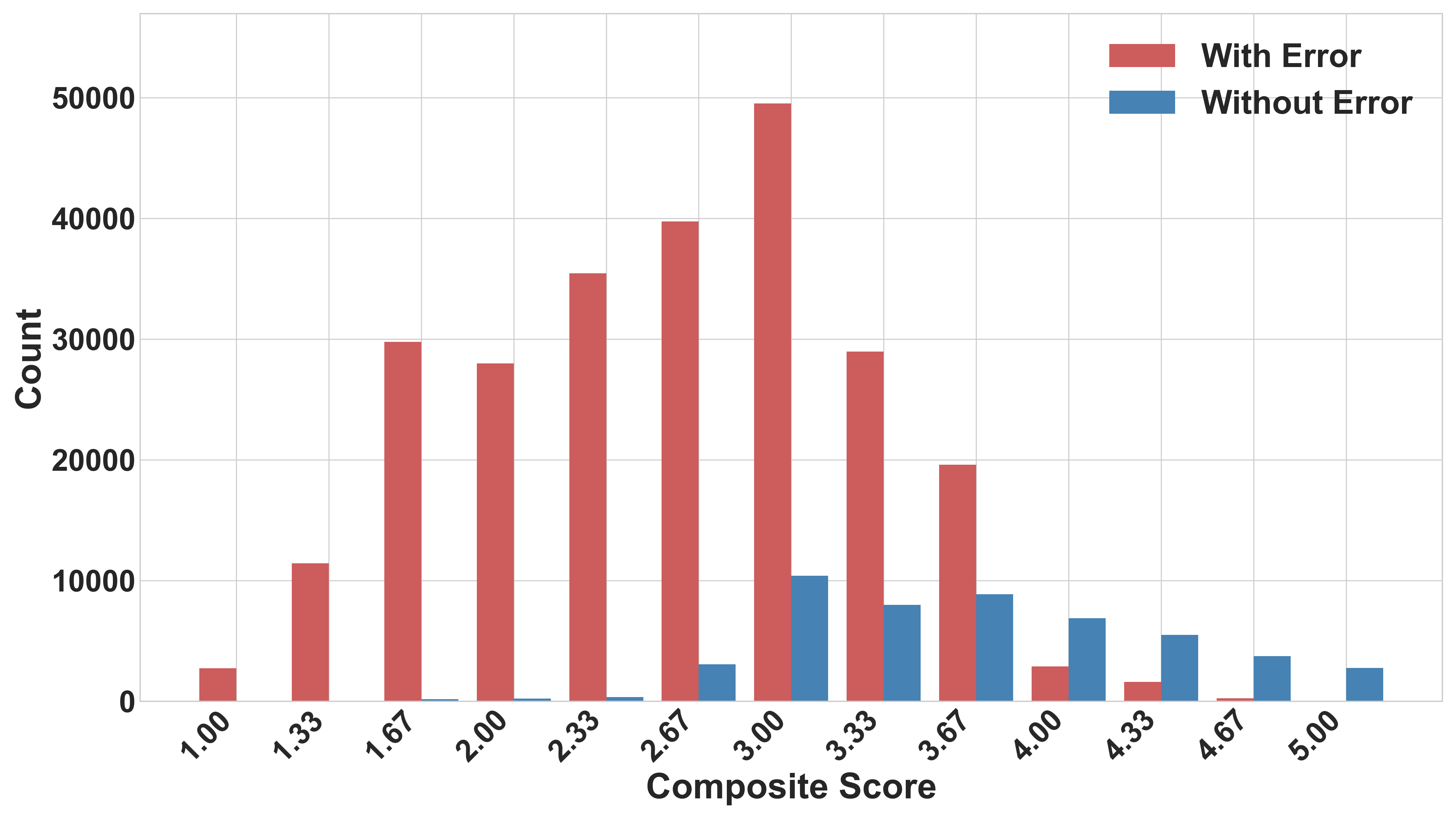} 
    \caption{
    Score distribution comparing original and defect-injected samples, illustrating how EVIAN separates high-quality data from subtle semantic corruptions.
    }
    \label{fig:score_distribution}
\end{figure}

To evaluate EVIAN's discriminative power, we apply it to our benchmark containing 50,000 pristine and 250,000 defect-injected samples. As illustrated in Figure~\ref{fig:score_distribution}, the two groups display a clear separation: 92.3\% of pristine entries score $\ge$ 3.0, while defect-injected samples form a distinct mid-range peak around 3.0. This shift indicates that EVIAN effectively penalizes semantically corrupted responses and differentiates them from high-quality data.

The separation is further quantified by a Jensen–Shannon divergence of 0.35 and an AUC of 0.86, demonstrating the metric’s strong discriminative capability. Furthermore, the concentration of defective samples in the mid-range, instead of accumulating at the lowest scores, shows that EVIAN detects subtle, context-dependent defects such as logical fallacies, not merely coarse inconsistencies. Although a small fraction of injected samples retain higher scores due to nuanced semantic ambiguities, the pronounced distributional gap overall provides strong evidence that EVIAN functions as a robust, fine-grained filter for high-quality data curation.

\subsection{Downstream Task Performance}
\label{sec:downstream_performance}
To evaluate the practical impact of EVIAN, we fine-tuned models on 10K-sample subsets curated by different methods and compared their downstream performance across multiple benchmarks. As shown in Table~\ref{tab:main_results}, \textbf{the model trained on the EVIAN-selected subset achieves the current best performance} (average score of 70.20), surpassing both the previous SOTA method (SCALE, 67.41) and the model trained on the full 300K unfiltered dataset (63.77). This ``less is more'' result highlights the diagnostic precision of EVIAN, which consistently extracts higher-quality data from a noisy pool.

These gains stem from EVIAN’s ``Decomposition-then-Evaluation'' paradigm, which addresses fine-grained defects overlooked by coarse auditing approaches. Filters such as CLIPScore and BLIP-2 provide moderate improvements but fail to capture errors like factual inaccuracies or logical fallacies. Moreover, EVIAN outperforms the \textbf{Qwen2.5-VL baseline} (70.20 vs. 66.34) using the identical auditor architecture, suggesting that the performance gains stem from our structured verification logic rather than simple knowledge distillation. By explicitly evaluating Image-Text Consistency, Logical Coherence, and Factual Accuracy, EVIAN yields targeted diagnostics that translate into stronger downstream models. For example, EVIAN’s leading performance on MME (1876.89) and POPE (79.87) validates its ability to mitigate hallucinations through holistic multi-dimensional verification, while its gains on A-OKVQA (0.7493) and ScienceQA (0.7115) highlight the benefit of auditing factual and reasoning components.

\begin{table*}[t]
\caption{Comparisons with state-of-the-art data selection baselines on 10K subsets, where \textbf{bold} and \underline{underlined} indicate the best and second-best results, respectively. ``Full Data'' denotes training on the entire 300K pool.}

\label{tab:main_results}
\centering
\resizebox{0.95\textwidth}{!}{%
\begin{tabular}{l|ccc|cc|c|c} 
\toprule
\textbf{Model} & \textbf{MME} & \textbf{MMBench} & \textbf{SEEDBench} & \textbf{ScienceQA} & \textbf{A-OKVQA} & \textbf{POPE} & \textbf{Avg} \\
\midrule
Random         & 1475.76          & 0.5353          & 0.6031          & 0.6614          & 0.7092          & 75.50          & 63.18 \\
Full Data      & 1553.05          & 0.5953          & 0.5743          & 0.6267          & 0.6934          & 78.17          & 63.77 \\
\midrule
CLIPScore      & 1565.29          & 0.5746          & 0.6170          & 0.6906          & \underline{0.7301}    & 74.57          & 65.28 \\
ALBEF          & 1590.70          & 0.6003          & 0.6107          & 0.6748          & 0.7048          & 72.29          & 64.69 \\
BLIP           & 1686.62          & 0.6183          & 0.6115          & 0.6802          & 0.6978          & 73.40          & 65.74 \\
BLIP-2         & 1810.34          & 0.6317          & 0.6187          & \underline{0.7045}    & 0.7127          & 77.38          & \underline{68.13} \\
SCALE          & \underline{1814.97}    & \underline{0.6318}    & \underline{0.6280}    & 0.6916          & 0.7066          & 73.81          & 67.41 \\
Qwen2.5-VL     & 1682.78          & 0.5796          & 0.6182          & 0.6797          & 0.7187          & \underline{78.30}    & 66.34 \\
\midrule
\rowcolor{mygray}
\textbf{EVIAN (Ours)} & \textbf{1876.89} & \textbf{0.6463} & \textbf{0.6359} & \textbf{0.7115} & \textbf{0.7493} & \textbf{79.87} & \textbf{70.20} \\
\bottomrule
\end{tabular}%
}
\end{table*}

Overall, these results reveal a fundamental limitation of existing curation strategies: high similarity or holistic scores do not guarantee utility and frequently obscure critical semantic defects. In contrast, EVIAN's multi-dimensional auditing yields cleaner and more reliable training data, enabling models trained on small, high-quality subsets to surpass those trained on far larger but noisier datasets. This points to a clear direction for LVLM development: progress hinges less on scaling data volume and more on fine-grained, interpretable auditing that enforces visual fidelity, factual accuracy, and logical coherence. To further verify that these gains arise from intrinsic data quality rather than a potential inductive-bias alignment between the Qwen-based auditor and target model, we conduct a cross-architecture evaluation with InternVL2-2B \citep{chen2024internvl}. As shown in Appendix~\ref{app:internvl_experiments}, EVIAN's advantages extend to a model family with a distinct architectural lineage, reinforcing the generality of our conclusions. Finally, we validate EVIAN on the original, unmodified data distribution in Appendix~\ref{app:real_world_effectiveness}, where a 10K curated subset rivals the full 300K baseline, confirming its capability to capture intrinsic data quality signals beyond artificial defects.

\subsection{Ablation Experiment}
\label{sec:ablation_experiment}

To validate the necessity of each component in EVIAN, we conducted an ablation study by selectively excluding the Logical Coherence ($S_L$), Factual Accuracy ($S_K$), and Image-Text Consistency ($S_V$) scores from the selection criteria. The results, summarized in Table~\ref{tab:ablation}, demonstrate that the full framework consistently achieves the best performance across all benchmarks. This confirms that these dimensions function as distinct and complementary factors essential for effective data auditing.

\begin{table*}[h]
\centering
\caption{Ablation study of the EVIAN framework, evaluating the contribution of individual components by removing them from the full pipeline. ``w/o Decomposition'' denotes the variant without the fine-grained decomposition stage. $S_L$ and $S_K$ represent the \textbf{Logical Coherence} and \textbf{Factual Accuracy} scores, respectively.
}
\label{tab:ablation}
\resizebox{0.95\textwidth}{!}{%
\begin{tabular}{l|ccc|cc|c|c}
\toprule
\textbf{Configuration} & \textbf{MME} & \textbf{MMBench} & \textbf{SEEDBench} & \textbf{ScienceQA} & \textbf{A-OKVQA} & \textbf{POPE} & \textbf{Avg} \\
\midrule
w/o Decomposition     & 1706.70 & 0.6401 & 0.6312 & 0.7085 & 0.7170 & 76.93 & 67.93 \\
w/o $S_L$             & 1656.62 & 0.3425 & 0.5324 & 0.5563 & 0.6288 & 78.45 & 57.27 \\
w/o $S_K$             & 1604.91 & 0.6110 & 0.5875 & 0.6604 & 0.6629 & 75.77 & 64.21 \\
w/o $S_L, S_K$ (Only $S_V$) & 1807.13 & 0.5605 & 0.6092 & 0.6822 & 0.7389 & 68.56 & 65.36 \\
\midrule
\rowcolor{mygray}
\textbf{EVIAN (Full)} & \textbf{1876.89} & \textbf{0.6463} & \textbf{0.6359} & \textbf{0.7115} & \textbf{0.7493} & \textbf{79.87} & \textbf{70.20} \\
\bottomrule
\end{tabular}%
}
\end{table*}

The full EVIAN framework achieves the best average performance (70.20), confirming the synergistic benefit of combining all three evaluation axes. Most notably, removing Logical Coherence ($S_L$) leads to a significant decline in performance to 57.27. This counterintuitive outcome arises because filtering based solely on factual accuracy ($S_K$) and visual consistency ($S_V$) inadvertently favors responses that are factually correct and visually grounded but logically inconsistent. The prevalence of such inconsistent samples introduces conflicting supervision signals, which significantly impairs the model's performance on reasoning-intensive benchmarks like ScienceQA. This confirms that logical integrity is not merely an auxiliary metric but a critical factor for complex reasoning tasks.

In contrast, relying solely on Image-Text Consistency ($S_V$) presents an inverse trade-off. While it yields a stable average score (65.36) by avoiding logical traps, it exhibits a marked drop on POPE (68.56), falling well below the random baseline. This sharp decline reveals that visual consistency alone is an insufficient proxy for data quality; without the structural constraints of logical and factual verification, the curation process fails to filter out subtle hallucinations, leaving the model vulnerable to object fabrication. Thus, EVIAN’s comprehensive auditing is essential: $S_V$ ensures visual relevance, while $S_L$ and $S_K$ are essential for guaranteeing logical coherence and factual precision.

\section{Conclusion}
\label{sec:conclusion}
In this work, our proposed visual instruction tuning data auditing method \textbf{EVIAN}, advances LVLM data quality auditing through three contributions: a 300K-sample benchmark with systematically injected defects, a ``Decomposition-then-Evaluation'' paradigm that separates visual, inferential, and factual components, and the EVIAN framework, which scores data along Image-Text Consistency, Logical Coherence, and Factual Accuracy. Experiments show that EVIAN-curated subsets consistently outperform models trained on much larger unfiltered datasets, and ablations confirm the necessity of each evaluation dimension. Surprisingly, our study also reveals that \textbf{dividing complex auditing into verifiable subtasks enables robust curation}, and that \textbf{Logical Coherence is the most critical factor for downstream reliability}. These results establish interpretable, fine-grained auditing as the foundation for advancing LVLMs.

\section{Acknowledgments}
Zimu Jia, Mingjie Xu, and Jiaheng Wei are partially supported by the CNPC Technology Project ``Research on Key Technologies of Artificial Intelligence for Oil and Gas Exploration and Development'' (No. 2023DJ84), and the Guangdong Provincial Key Laboratory of Integrated Communication, Sensing and Computation for Ubiquitous Internet of Things (No. 2023B1212010007).

\section{Limitations}
\label{sec:limitations}

While EVIAN demonstrates strong effectiveness for fine-grained auditing of visual instruction data, several limitations remain.
First, the framework relies on large pre-trained multimodal language models for both response decomposition and quality evaluation. Despite strong alignment between automated scores and human judgments, the auditing process may still inherit biases, blind spots, or reasoning tendencies from the underlying models, particularly in ambiguous or culturally sensitive scenarios.

Second, EVIAN assumes that model responses can be reliably decomposed into visual descriptions, subjective inferences, and factual claims. Errors introduced during this decomposition stage may propagate to subsequent evaluations and affect the final quality scores. Although our ablation results suggest that explicit decomposition is beneficial overall, improving robustness under imperfect decomposition remains an open challenge.

Third, the current pipeline incurs non-trivial computational cost due to multiple invocations of large models, which may limit its applicability in resource-constrained settings or when auditing extremely large-scale datasets. Exploring lighter-weight auditors or partially learned approximations of individual components is a promising direction for future work.

Finally, EVIAN focuses on auditing response quality with respect to visual grounding, logical coherence, and factual accuracy. Other important aspects, such as stylistic diversity, pedagogical value, or downstream task-specific preferences, are not explicitly modeled and may require complementary criteria depending on the application context.

\section{Ethical Considerations}
\label{sec:ethics}

This work aims to improve the quality and reliability of visual instruction-tuning data through automated auditing. By identifying logical inconsistencies, factual errors, and visual misalignment in training data, EVIAN has the potential to reduce hallucinations and misleading outputs in downstream vision--language models, contributing to safer and more trustworthy AI systems.

Nevertheless, the use of large language and vision--language models as automated auditors raises ethical considerations. Model-based judgments may reflect biases present in their training data and influence which data distributions are preserved or suppressed during curation, potentially leading to systematic over- or under-filtering of certain types of content. Care should therefore be taken when deploying EVIAN in sensitive domains, such as medical or legal applications, where incorrect filtering decisions may have disproportionate consequences. EVIAN’s outputs should be viewed as decision-support signals rather than absolute ground truth, and human oversight remains important in high-stakes settings.

Finally, all experiments in this work are conducted on publicly available datasets, and no new human annotations are collected. The proposed framework is intended to support responsible data curation practices and does not introduce additional privacy or data collection concerns beyond those already present in existing vision--language datasets.

\bibliography{custom}

\newpage
\appendix
\section{Evian Framework Implementation Details}
\label{app:implementation_details}

\subsection{Models and Computational Resources}
All experiments were conducted on a high-performance computing node equipped with eight NVIDIA H100 (80GB) GPUs. We employed the Qwen3-235B-A22B-Instruct-2507-FP8 model for the text-heavy response decomposition and defect injection phases, and the \textbf{Qwen2.5-VL-7B-Instruct-AWQ} model for the multi-faceted quality assessment. Both models were deployed via vLLM (v0.10.0) using a greedy sampling strategy to ensure deterministic reproducibility within a unified software environment consisting of PyTorch (v2.7.1) and CUDA (v12.6). Leveraging this setup, the entire processing pipeline for the 300,000-sample benchmark was completed in nearly 28 hours.

\subsection{Supervised Fine-Tuning (SFT) Details}
To efficiently fine-tune the \textbf{Qwen2-VL-2B} base model, we implemented a selective update strategy, freezing the vision tower while training the projector MLP and language model. This training process was conducted on a server equipped with eight NVIDIA vGPU (48 GB) cards. It leveraged DeepSpeed ZeRO Stage 3 for memory optimization, resulting in an effective global batch size of 128. All key hyperparameters are detailed in Table~\ref{tab:sft_hyperparams}.

\begin{table}[h]
\caption{Supervised Fine-Tuning (SFT) Hyperparameters for the Base Model.}
\label{tab:sft_hyperparams}
\begin{center}
\begin{tabular}{@{}ll@{}}
\toprule
\textbf{Hyperparameter} & \textbf{Value} \\ \midrule
Base Model & Qwen2VL-2B \\
Epochs & 1 \\
Learning Rate & $5 \times 10^{-6}$ \\
Batch Size (per device) & 2 \\
Gradient Accumulation Steps & 8 \\
Weight Decay & 0.0 \\
Warmup Ratio & 0.1 \\
LR Scheduler & Cosine \\
Max Gradient Norm & 1.0 \\
Precision & BF16 \\
Max Sequence Length & 8192 \\
Gradient Checkpointing & Enabled \\
Optimization & ZeRO-3 \\
\bottomrule
\end{tabular}
\end{center}
\end{table}

\subsection{Prompt Engineering for Phase 1: Response Decomposition}

\begin{promptdisplay}{Step 1: Prompt for Semantic Tagging}
Response: \{response\}

Your task is to precisely insert \textless INFER\textgreater~for subjective judgments and \textless KNOW\textgreater~for external knowledge.

\textbf{Critical Guidelines for Annotation:}
\begin{enumerate}
    \item \textbf{Tag the Complete Thought:} Precisely wrap the shortest, complete phrase that conveys the entire logical idea (like a cause-and-effect statement) or the full piece of external information.
    \item \textbf{Tag Interpretations of Effect/Cause:} Always tag phrases that describe the effect, purpose, or reason for a visual element.
    \item \textbf{Strictly Visual is NOT Tagged:} DO NOT tag objective, verifiable descriptions of visual facts.
    \item \textbf{Do Not Change Words:} Do not add, delete, or rephrase any original words, like Visible Text or Numbers.
    \item \textbf{Output Format:} Your response must start with the prefix ``Marked Response:''.
\end{enumerate}

\textbf{Examples:}

\textbf{Input:} The lighting in the room is soft, creating a cozy atmosphere. The design suggests it is from the Victorian era. \par
\textbf{Output:} Marked Response: The lighting in the room is soft, \textless INFER\textgreater creating a cozy atmosphere\textless/INFER\textgreater. \textless INFER\textgreater The design suggests it is from the Victorian era\textless/INFER\textgreater.

\textbf{Input:} This is a 1976 postage stamp from Hungary, a country in Central Europe. \par
\textbf{Output:} Marked Response: This is a 1976 postage stamp from Hungary, \textless KNOW\textgreater a country in Central Europe\textless/KNOW\textgreater.

\textbf{Input:} The image shows a can of Coca-Cola. \par
\textbf{Output:} Marked Response: The image shows a can of Coca-Cola.
\end{promptdisplay}

\begin{promptdisplay}{Step 2: Prompt for Visual Distillation}
Instruction: \{instruction\} \par
Annotated Response: \{marked\_response\}

Task: Process the ``Annotated Response'' by modifying ONLY the segments wrapped in \textless INFER\textgreater...\textless/INFER\textgreater~or \textless KNOW\textgreater...\textless/KNOW\textgreater~tags.
\begin{itemize}
    \item Rewrite or entirely remove tagged segments to leave only what is directly and objectively visible in the image.
    \item \textbf{Crucially, all content NOT wrapped in tags MUST be preserved exactly as is, without any modification.}
\end{itemize}

\textbf{Guidelines:}
\begin{enumerate}
    \item \textbf{Rewrite When Possible:} If a tagged idea can be rephrased as a neutral, objective, image-based description, rewrite it and remove the tags. For example, change ``\textless INFER\textgreater creating a cozy atmosphere\textless/INFER\textgreater'' to ``which illuminates the scene.''
    \item \textbf{Delete When Necessary:} For clearly irrelevant or purely speculative content that cannot be visually confirmed, delete the entire tagged segment (including the tags).
    \item \textbf{No New Information:} DO NOT introduce any new guesses, opinions, or visual details that were not already present in the untagged parts of the original response.
    \item \textbf{Output Format:} Your response must start with the prefix ``Cleaned Response:''.
\end{enumerate}

\textbf{Example:} \par
Input Annotated Response: \par
A person wearing sunglasses stands under a tree. \textless INFER\textgreater She must be shielding her eyes from harsh sunlight.\textless/INFER\textgreater~Leaves are scattered on the ground. \textless KNOW\textgreater This park is famous for its autumn foliage tours.\textless/KNOW\textgreater

Output: \par
Cleaned Response: A person wearing sunglasses stands under a tree. Leaves are scattered on the ground.
\end{promptdisplay}

\begin{promptdisplay}{Step 3: Prompt for Fluent Synthesis}
Instruction: \{instruction\} \par
Cleaned Response: \{cleaned\_response\}

Task: Rephrase the ``Cleaned Response'' into a single, cohesive, and purely visual description.

\textbf{Guidelines:}
\begin{enumerate}
    \item \textbf{Strictly Adhere to Input:} Your output MUST be a faithful reorganization of ONLY the information present in the ``Cleaned Response.''
    \item \textbf{Preserve All Details:} Do not omit any visual information. Every object, attribute, and spatial relation from the input must be represented in your summary.
    \item \textbf{No New Content or Inference:} Crucially, DO NOT add any new visual details, reasoning, assumptions, or subjective/interpretive language (e.g., ``beautiful'', ``seems like'', ``creates a sense of''). Your job is to describe, not to analyze.
    \item \textbf{Improve Flow:} Focus on improving sentence structure and grammatical correctness to create a natural-sounding paragraph.
    \item \textbf{Output Format:} Your response must start with the prefix ``Visual Summary:''.
\end{enumerate}

\textbf{Example:} \par
Input Cleaned Response: A white cat is on a windowsill. The background shows buildings. Light is coming through the window. \par
Output: \par
Visual Summary: A white cat sits on a windowsill where bright light is streaming in. Buildings are visible in the background.
\end{promptdisplay}

\subsection{Prompting and Rubrics for Phase 2: Multi-faceted Quality Assessment}

\begin{promptdisplay}{Dimension $S_L$: Prompt for Logical Coherence}
Input Text for Evaluation: \{text\_to\_evaluate\}

Task: You are an AI assistant designed to evaluate the correctness of logical reasoning. Your primary focus is to rigorously scrutinize the logical soundness and validity of the reasoning contained ONLY within the \textless INFER\textgreater...\textless/INFER\textgreater~tags, based on the visual evidence in the image.

\textbf{Evaluation and Scoring Rules:}
\begin{enumerate}
    \item Isolate and Evaluate: Focus exclusively on the statements inside the \textless INFER\textgreater~tags.
    \item Assess Plausibility against Image: Judge if the inference is a logical and plausible conclusion derived from the visual information in the image.
    \item Output Format:
        \begin{itemize}
            \item Score: integer 1-5
            \item Explanation: A brief evaluation of the logical rigor, noting key flaws or strengths.
        \end{itemize}
\end{enumerate}

\textbf{Scoring Rubric:}
\begin{description}
    \item[Score 1:] Grossly Illogical or Baseless. The inference is pure speculation with no connection to the image (e.g., predicting the future from a photo of a cat), or it's self-contradictory.
    \item[Score 2:] Significant Logical Gaps. The inference is a major leap in logic. While loosely related to the image, it is highly unlikely or requires many unsupported assumptions. (e.g., ``A person is running, \textless INFER\textgreater so this must be a professional athlete training for the Olympics\textless/INFER\textgreater.'' )
    \item[Score 3:] Plausible but Unprovable. The inference is reasonable and could be true, but it is not strongly supported by visual evidence and remains a subjective interpretation. (e.g., ``The room is dim, \textless INFER\textgreater creating a sad atmosphere\textless/INFER\textgreater.'' )
    \item[Score 4:] Logically Sound. The inference is very likely correct and follows directly from strong visual evidence, with only very minor room for doubt. (e.g., ``The man holds an umbrella, \textless INFER\textgreater suggesting it is raining or about to rain\textless/INFER\textgreater.'' )
    \item[Score 5:] Logically Airtight. The inference is an undeniable conclusion based on the visual facts and common-sense logic; it is virtually irrefutable. (e.g., ``The wreck shows a crushed car, \textless INFER\textgreater indicating a high-impact collision occurred\textless/INFER\textgreater.'' )
\end{description}
\end{promptdisplay}

\begin{promptdisplay}{Dimension $S_K$: Prompt for Factual Accuracy}
Input Text for Evaluation: \{text\_to\_evaluate\}

Task: You are an expert fact-checking assistant. Your task is to evaluate the factual correctness of the information contained ONLY within the \textless KNOW\textgreater...\textless/KNOW\textgreater~tags. Base your assessment on your internal, general knowledge.

\textbf{Output Format:} \par
Score: integer 1-5 \par
Explanation: A brief justification for your score, specifying which facts are correct or incorrect.

\textbf{Scoring Rubric:}
\begin{description}
    \item[Score 1: \textbf{Entirely Incorrect or Fabricated.}] The information is factually wrong, nonsensical, or a complete fabrication (e.g., contains imaginary objects like the `Luminara Scepter').
    \item[Score 2: \textbf{Largely Incorrect.}] Contains a core factual error, even if minor details are correct. (e.g., ``\textless KNOW\textgreater Paris, the capital of England...\textless/KNOW\textgreater''). The presence of a single major error means the score cannot be higher than 2.
    \item[Score 3: \textbf{Partially Correct but Misleading.}] Contains a mix of correct and incorrect information, or the information is technically correct but presented in a highly misleading context.
    \item[Score 4: \textbf{Mostly Correct.}] The core assertion is factually sound but contains a minor, non-critical inaccuracy (e.g., a slightly wrong year, a minor detail about a standard feature).
    \item[Score 5: \textbf{Fully Correct and Accurate.}] Every single claim within the tags is factually sound, precise, and widely accepted.
\end{description}
\end{promptdisplay}

\begin{promptdisplay}{Dimension $S_V$: Prompt for Image-Text Consistency}
Input Text: \{text\_input\}

Task: You are a visual consistency scoring assistant. Your task is to evaluate whether the extracted text description’s assertions can be verified by the given image. Only assess consistency, not completeness: do NOT penalize the description for omitting image details, but DO penalize any assertions that contradict or cannot be supported by the image.

\textbf{CORE SCORING GUIDELINE:} Be decisive in your scoring. If the description is fully and accurately supported by the image without any errors, the score must be 5. Do not default to 4 if a 5 is warranted.

\textbf{Output Format:} \par
Score: integer 1-5 \par
Explanation: Brief justification, indicating which assertions are verifiable and which are inconsistent or unclear.

\textbf{Scoring Rubric:}
\begin{description}
    \item[Score 1:] Severely inconsistent or completely unrelated. Most or all assertions contradict the image.
    \item[Score 2:] Largely inconsistent. Only one or two minor assertions can be matched to the image.
    \item[Score 3:] Partially consistent. Some key assertions align with the image, but others are vague, potentially incorrect, or unsupported.
    \item[Score 4:] Mostly consistent. The bulk of assertions are supported by the image, but there is at least one minor imprecision or slight unsupported detail that does not mislead. Use this score for responses that are good but not perfect.
    \item[Score 5:] Fully consistent and accurate. Every single assertion in the text is clearly and precisely verifiable in the image. There are no unsupported or contradictory claims. If all claims are verified, you MUST assign this score.
\end{description}
\end{promptdisplay}

\section{Defect Injection Pipeline and Prompt Catalog}
\label{app:dataset_construction}

To create a challenging and diverse evaluation set, we designed and implemented a three-stage, LLM-driven pipeline for injecting controlled, contextually-relevant defects into high-quality responses. This automated pipeline ensures that the generated errors are not random but are intelligently tailored to the content of the source text.

\subsection{The Three-Stage Defect Injection Pipeline}
The core of our data generation process is a sequential pipeline that first analyzes the text, then selects an appropriate error type, and finally rewrites the text to introduce the defect.

\paragraph{Stage 1: Content Analysis} First, an LLM analyzes the source text to determine if it contains logical reasoning or external knowledge. This classification serves as a prior for the subsequent error selection stage. The analysis is performed using the prompt below.

\begin{promptdisplay}{Prompt for Content Analysis}
You are a text analysis expert. Analyze the following text and
determine if it contains a) logical reasoning, inference, or
conclusion, and b) specific external knowledge (like names of people,
places, brands, historical facts).

Respond ONLY with a JSON object with two boolean keys:

\{``contains\_reasoning'': boolean, ``contains\_knowledge'': boolean\}.

Text to analyze:
``\{text\_to\_analyze\}''
\end{promptdisplay}

\paragraph{Stage 2: Category and Subtype Selection} The primary error category is selected via a probabilistic cascade that prioritizes the knowledge category with a probability of 0.8 for texts flagged contains\_knowledge, followed by the reasoning category with a probability of 0.6 for those with contains\_reasoning, and otherwise defaults to the consistency category. This initial choice, in turn, dictates the method for subtype determination: while subtypes for the consistency category are chosen uniformly at random, a more nuanced approach is employed for the contextually-sensitive knowledge and reasoning categories, for which a second LLM call intelligently selects the most plausible subtype using the following prompt.

\begin{promptdisplay}{Prompt for Category and Subtype Selection}
You are a text analysis expert. Your task is to select the single best error-injection strategy for the ``Original Text'' from the ``Available Options''.

\textbf{Available Options:}
\{error\_options\_text\}

\textbf{Original Text:}
``\{text\_to\_analyze\}''

Analyze the text and choose the error code from the options that is most relevant to the text's content. Respond ONLY with a JSON object containing your choice.
\end{promptdisplay}

\paragraph{Stage 3: Defect Generation} Finally, with a specific error subtype selected, a third LLM call rewrites the original text according to the corresponding instruction. The final prompt is constructed from a template, and a strict system prompt is used to ensure clean output.

\begin{promptdisplay}{Prompts for Defect Generation}
\textbf{Instruction:} \{prompt\_instruction\} \par
\textbf{Original Text to Corrupt:} ``\{original\_text\}'' \par
- - - \par
You MUST provide ONLY the corrupted/rewritten text as the output. Do not include any preambles, explanations, or wrappers like `Rewritten Text:' or the original response in your final output.
\end{promptdisplay}

\subsection{Catalog of Defect Injection Instructions} The complete set of instructions used in the defect generation stage is detailed below.

\paragraph{A. Consistency Errors}
\begin{itemize}
    \item consistency\_attribute: Rewrite the response by changing an attribute (like color, count, or size) of one key object.
    \item consistency\_spatial: Rewrite the response by incorrectly describing the spatial relationship between two objects (e.g., change `on the table' to `under the table').
    \item consistency\_action: Rewrite the response by describing an incorrect action or state for a subject (e.g., change `a man is sitting' to `a man is running').
    \item consistency\_fake: Rewrite the response to include a mention of a plausible but non-existent object.
    \item consistency\_misidentification: Rewrite the response by misidentifying an existing object (e.g., call a `cup' a `bowl').
\end{itemize}

\paragraph{B. Reasoning Correctness Errors}
\begin{itemize}
    \item reasoning\_conclusion: Your task is to rewrite the text by making a hasty generalization. The method is to grab a single detail from the text (such as one person running) and then extrapolate it into a grand conclusion that seems plausible but is actually very arbitrary (such as concluding this must be a professional marathon training session). Ensure you use reasoning words like `so' or `therefore' to connect this flawed logical chain.
    \item reasoning\_causal: Your task is to confuse correlation with causation. Find two things in the text that might happen concurrently but have no direct causal link, and then forcibly establish a cause-and-effect relationship between them using words like `because' or `leading to'. For instance, you could take the action `a man holding an umbrella indoors' and incorrectly present it as the cause for `a power outage in the room', creating a deceptive misattribution.
    \item reasoning\_prediction: Your task is to make an overly arbitrary and confident prediction based on extremely limited information. You need to take a trivial, small action (such as a child stacking blocks) and lead it directly to a very grand and distant future (such as predicting they will surely become a great architect). This prediction needs to sound physically possible, but its logical leap must be huge and baseless.
    \item reasoning\_procedural: Your task is to, within a normal process description, insert a step that seems plausible but is actually superfluous or based on pseudoscience. This step must not cause the entire process to fail but will make it logically flawed. For instance, when describing the process of brewing tea, you could add a step claiming that `before adding water, you need to let the tea leaves sit for a minute to absorb the room's energy,' thereby making the process imprecise.
    \item reasoning\_comparison: Your task is to construct a faulty analogy. You need to find two things that have only minor superficial similarities but are completely different in their core essence to make a comparison, and then draw a misleading conclusion from it. A classic example is to compare `company strategy' to a `car engine' and then argue that `as long as there's enough fuel (funding), success is guaranteed,' an analogy that deliberately ignores more critical factors like the `steering wheel (strategic direction)'
\end{itemize}

\paragraph{C. External Knowledge Errors}
\begin{itemize}
    \item knowledge\_entity: If the response mentions a real-world named entity, rewrite it by corrupting that entity (e.g., `Eiffel Tower in London').
    \item knowledge\_context: Rewrite the response to place an object or scene in a wrong historical or technological context.
    \item knowledge\_definition: If the response defines a concept, rewrite it to provide an incorrect definition.
    \item knowledge\_attribution: If the response mentions a creation or quote, misattribute it to the wrong source.
\end{itemize}

\section{Sensitivity Analysis on Component Weights}
\label{app:sensitivity}

To further validate the robustness of EVIAN and explore the impact of different quality dimensions on downstream performance, we conducted a sensitivity analysis by adjusting the aggregation weights refer to Section~\ref{sec:ranking_selection}.

We designed three biased weighting schemes, where one dimension is assigned a dominant weight of 60\% ($w=0.6$), while the remaining two dimensions are assigned 20\% ($w=0.2$) each. The configurations are:
\begin{itemize}
    \item \textbf{Vision-Centric:} $w_V=0.6, w_L=0.2, w_K=0.2$. Prioritizes visual fidelity.
    \item \textbf{Reason-Centric:} $w_L=0.6, w_V=0.2, w_K=0.2$. Prioritizes logical soundness.
    \item \textbf{Knowledge-Centric:} $w_K=0.6, w_V=0.2, w_L=0.2$. Prioritizes factual correctness.
\end{itemize}

We selected the top-10k samples using each scheme and fine-tuned the model under the same protocol as our main experiments. The results are reported in Table~\ref{tab:sensitivity_results}.

\begin{table*}[h]
\caption{Sensitivity analysis of EVIAN under different weighting schemes. We report performance on 10K-sample subsets selected by prioritizing different quality dimensions (60\% weight). \textbf{Reason-Centric} selection achieves the best average performance, highlighting the importance of logical coherence. \textbf{Bold} denotes the best result, and \underline{underlined} denotes the second best.}
\label{tab:sensitivity_results}
\centering
\resizebox{\textwidth}{!}{%
\begin{tabular}{l|ccc|cc|c|c}
\toprule
\textbf{Configuration} & \textbf{MME} & \textbf{MMBench} & \textbf{SEEDBench} & \textbf{ScienceQA} & \textbf{A-OKVQA} & \textbf{POPE} & \textbf{Avg} \\
\midrule
Vision-Centric ($S_V \uparrow$) & \underline{1779.15} & \underline{0.6109} & \underline{0.6163} & \underline{0.6905} & \underline{0.7160} & \textbf{77.76} & \underline{67.45} \\
Reason-Centric ($S_L \uparrow$) & \textbf{1803.53} & \textbf{0.6373} & \textbf{0.6312} & \textbf{0.7030} & \textbf{0.7228} & \underline{75.81} & \textbf{68.28} \\
Know-Centric ($S_K \uparrow$)   & 1748.02 & 0.5995 & 0.6034 & 0.6731 & 0.7021 & 75.74 & 65.99 \\
\bottomrule
\end{tabular}%
}
\end{table*}

The results suggest that emphasizing logical coherence is particularly beneficial under the evaluated setting, as the \textit{Reason-Centric} configuration attains the highest average score (68.28) and shows consistent advantages on reasoning-intensive benchmarks. In contrast, the \textit{Vision-Centric} scheme demonstrates strong effectiveness in reducing hallucinations, as reflected by its POPE performance, but does not consistently match the balanced EVIAN setting, indicating that visual accuracy alone is insufficient for broader capability gains. The \textit{Know-Centric} configuration yields comparatively lower performance (65.99), suggesting potential trade-offs between strict factual filtering and the retention of samples that support multi-step reasoning. Overall, these findings indicate that while balanced weighting provides a robust default, the EVIAN framework allows practitioners to adjust score weights to prioritize task-specific objectives, such as hallucination mitigation via $S_V$ or enhanced reasoning via $S_L$.

\section{Generalizability Across Model Architectures}
\label{app:internvl_experiments}

\paragraph{Motivation: Addressing Auditor--Target Architectural Bias.}
A potential concern regarding our main results is the possibility that EVIAN overfits to the inductive biases shared by the auditor (Qwen2.5-VL-7B) and the fine-tuned target model (Qwen2-VL-2B). Since these two models belong to the same family, one could argue that EVIAN does not truly identify intrinsically high-quality samples, but instead selects data that align with Qwen-specific reasoning heuristics. This concern is particularly acute for the \textbf{Logical Coherence} dimension, which depends on the auditor’s internal reasoning pathways and, unlike Visual Consistency, could in principle be architecture-specific. To rigorously stress-test this hypothesis, we adopt \textbf{InternVL2-2B} \citep{chen2024internvl} as an alternative target model. InternVL2 differs substantially from Qwen in both its vision encoder and language backbone, while maintaining a comparable 2B scale, allowing us to isolate architectural effects from capacity effects.

\paragraph{Experimental Setup.}
We follow the exact protocol of Section~\ref{sec:experiments} and fine-tune InternVL2-2B using 10k-sample subsets selected by three representative methods: Random Sampling, BLIP-2, and EVIAN. This setup ensures that the only change relative to the main experiment is the replacement of the downstream architecture. Any shift or preservation of performance patterns therefore directly reflects whether EVIAN captures genuinely architecture-agnostic indicators of data quality, or merely capitalizes on Qwen-specific inductive biases.

\paragraph{Results and Interpretation.}
The results are presented in Table~\ref{tab:internvl_results}. Across all benchmarks, the performance ordering observed in our main experiments, \textbf{EVIAN $>$ BLIP-2 $>$ Random}, remains fully preserved when InternVL2-2B is used as the downstream model. The consistency of this ranking across architectures strongly contradicts the architectural-bias hypothesis and indicates that EVIAN captures transferable signals of data quality rather than family-specific preferences. Notably, the largest gains again appear on reasoning-centric tasks such as \textbf{ScienceQA} (92.46 vs. 91.32) and \textbf{A-OKVQA} (76.24 vs. 74.67), demonstrating that the logical flaws identified by the Qwen-based auditor are also detrimental for a model with a completely different architecture. This shows that \textbf{logical coherence is a genuinely universal data-quality factor}, and that enforcing it yields robust improvements regardless of the model family used for fine-tuning.

\begin{table*}[h]
\caption{Cross-architecture validation using \textbf{InternVL2-2B}. We report the performance on 10K subsets. \textbf{Bold} denotes the best result, and \underline{underlined} denotes the second best. Note that the MME and MMBench columns have been mapped to their standard metric scales based on the raw logs. Our \textbf{EVIAN} consistently outperforms baselines even on a different model architecture.}
\label{tab:internvl_results}
\centering
\resizebox{\textwidth}{!}{%
\begin{tabular}{l|ccc|cc|c|c} 
\toprule
\textbf{Model} & \textbf{MME} & \textbf{MMBench} & \textbf{SEEDBench} & \textbf{ScienceQA} & \textbf{A-OKVQA} & \textbf{POPE} & \textbf{Avg} \\
\midrule
Random         & \underline{1776.87}    & 0.3857          & 0.6064          & 0.7313          & 0.5467          & \underline{87.18}    & 62.94 \\
BLIP-2         & 1754.31          & \underline{0.6945}    & \underline{0.6888}    & \underline{0.9132}    & \underline{0.7467}    & 87.10          & \underline{75.68} \\
\midrule
\rowcolor{mygray} 
\textbf{EVIAN (Ours)} & \textbf{1796.41} & \textbf{0.7096} & \textbf{0.6971} & \textbf{0.9246} & \textbf{0.7624} & \textbf{87.37} & \textbf{76.82} \\
\bottomrule
\end{tabular}%
}
\end{table*}

\begin{table*}[!htbp]
\caption{Comparison between the full original dataset and the EVIAN-selected subset. The ``Origin Full'' model is trained on the unmodified 300K source data. EVIAN achieves a higher average score using only 1/30th of the data, demonstrating superior data efficiency and quality verification.}
\label{tab:real_world_comparison}
\centering
\resizebox{\textwidth}{!}{%
\begin{tabular}{l|ccc|cc|c|c}
\toprule
\textbf{Model} & \textbf{MME} & \textbf{MMBench} & \textbf{SEEDBench} & \textbf{ScienceQA} & \textbf{A-OKVQA} & \textbf{POPE} & \textbf{Avg} \\
\midrule
Origin Full (300K) & 1715.03 & \textbf{0.6734} & \textbf{0.6416} & \textbf{0.7287} & \textbf{0.7511} & 79.68 & 70.07 \\
\midrule
\rowcolor{mygray}
\textbf{EVIAN (10K)} & \textbf{1876.89} & 0.6463 & 0.6359 & 0.7115 & 0.7493 & \textbf{79.87} & \textbf{70.20} \\
\bottomrule
\end{tabular}%
}
\end{table*}

\section{Effectiveness on Real-World Data}
\label{app:real_world_effectiveness}
To definitively address potential concerns regarding the generalizability of our method beyond the defect-injected benchmark, we conducted an additional evaluation on the original, unmodified 300K source dataset. This experiment aims to verify whether the data quality improvements observed with EVIAN stem solely from filtering out artificial noise or if the framework captures intrinsic data quality signals applicable to real-world scenarios. Our findings confirm the latter: the model fine-tuned on the small, EVIAN-curated subset outperforms the model trained on the entire original dataset, validating the efficacy of our pipeline in extracting high-value samples from standard distribution data.

\textbf{Experimental Setup.} We established a rigorous baseline using the ``Origin Full'' configuration, where the Qwen2-VL-2B model was fine-tuned on the complete pool of 300,000 raw samples derived from the eight foundational datasets prior to any defect injection. This represents the upper bound of data quantity without our intervention. This baseline was compared against the ``EVIAN (10K)'' model, which was fine-tuned on the top-10,000 samples selected by our framework. Both models utilized identical model architectures and training hyperparameters and were evaluated using the standard VLMEvalKit suite to ensure a fair comparison focused strictly on data efficiency and quality.

\textbf{Results.} As shown in Table~\ref{tab:real_world_comparison}, the EVIAN-selected subset achieves a higher average score of 70.20 compared to 70.07 for the full original dataset, despite utilizing only approximately 3.3\% of the total training volume. While the full dataset exhibits a slight advantage on knowledge-heavy benchmarks such as MMBench, SEEDBench, and ScienceQA, likely attributable to the broader exposure to diverse facts inherent in larger data scales, EVIAN secures a substantial lead on MME (+161.86) and maintains highly competitive performance on hallucination and reasoning tasks. This empirical evidence reinforces the ``less is more'' principle, suggesting that fine-grained auditing can effectively identify and prioritize high-utility samples even within generally high-quality data distributions.

\end{document}